\documentclass[twoside,11pt]{article}

\usepackage{jmlr2e}
\usepackage{multirow}
\usepackage{booktabs}
\usepackage{comment}

\ShortHeadings{End2You -- The Imperial e2e Toolkit}{Tzirakis, Zafeiriou, Schuller}
\firstpageno{1}

\begin{document}

\title{End2You -- The Imperial Toolkit for Multimodal Profiling by End-to-End Learning}

\author{\name Panagiotis Tzirakis \email panagiotis.tzirakis12@imperial.ac.uk 
       \AND
       \name Stefanos Zafeiriou \email s.zafeiriou@imperial.ac.uk 
       \AND
       \name Bj\"orn W. Schuller \email bjoern.schuller@imperial.ac.uk \\
       \addr Department of Computing\\
       Imperial College London\\
       London, UK }

\maketitle

\begin{abstract}

We introduce End2You -- the Imperial College London toolkit for multimodal profiling by end-to-end deep learning. End2You is an open-source toolkit implemented in Python and is based on Tensorflow. It provides capabilities to train and evaluate models in an end-to-end manner, i.\,e., using raw input. It supports input from raw audio, visual, physiological or other types of  information or combination of those, and the output can be of an arbitrary representation, for either classification or regression tasks. To our knowledge, this is the first toolkit that provides generic end-to-end learning for profiling capabilities in either unimodal or multimodal cases. To test our toolkit, we utilise the RECOLA database as was used in the AVEC 2016 challenge. Experimental results indicate that End2You can provide comparable results to state-of-the-art methods despite no need of expert-alike feature representations, but self-learning these from the data ``end to end''.

\end{abstract}

\begin{keywords}
end-to-end learning, multimodal profiling, multisensorial data, deep learning, affective computing
\end{keywords}

\section{Introduction}

One of the most important steps in the traditional machine learning pipeline is feature extraction. A number of hand-crafted features have been proposed throughout the years that have shown to be robust. Examples can be found in audio analysis where the Mel-Frequency Cepstral Coefficients (MFCC) (Logan (2000)), or the Perceptual Linear Prediction (PLP) coefficients  (Hermansky (1990)) are widely used, but also in computer vision with several popular features such as the Scale Invariant Feature Transform (SIFT) (Ng and Henikoff (2003))  and the Histogram of Oriented Gradients (HOG) (Dalal and Triggs (2005)).

However, in recent years, Deep Neural Networks (DNN) have gained much popularity due to repeatedly better performing, and often more robust features that can be extracted automatically, as compared to the hand-engineered ones. Additionally, suited networks can provide the capability to be trained in an end-to-end manner, i.\,e., using \textit{raw} information (Trigeorgis et al. (2016)), and the least possible human a-priori knowledge (Graves and Jaitly (2014)). To this end, several deep architectures have been proposed such as Convolutional Neural Networks (CNNs)  (Lecun (1989)) or Deep Belief Networks (DBNs) (Hinton et al. (2006)). To exploit the potential temporal information in the data, which is often of time series nature,  Recurrent Neural Networks (RNN) with memory have been proposed such as the Long Short-Term Memory (LSTM)  (Hochreiter and Schmidhuber (1997)) and the Gated Recurrent Unit (GRU) models  (Chung et al. (2014)).

Several areas of pattern recognition have benefited from these networks, such as the affective computing and computational paralinguistic domains -- e.\,g., for emotion recognition (Kim et al. (2013)). To further increase the performance of such systems, multiple inputs from different modalities, such as audio and video, are often considered  (Tzirakis et al. (2017)). In this paper and context, we propose the first of its kind End2You tool -- the Imperial College London toolkit for multimodal profiling. The toolkit provides models for training and evaluating with either unimodal or multimodal input. The input can be audio, video, audiovisual, or any other kind of modality or sensor data such as physiological data. In addition, the models can be easily combined in any desirable way. This elegantly includes fusion aspects of these heterogeneous information streams.

\section{Multimodal Profiling}

Our toolkit provides a set of models to extract information from raw data. Depending on the nature of the data, a number of models can be defined.

\textit{Audio}. This model processes the raw audio signal. It is comprised of a 2-block of convolution max-pooling layers. In the first block, the convolution layer has 40 filters of size 20 and the max-pooling layer has size 2. In the second block, the layer has 40 filters of size 40, and the pooling layer is applied to the feature maps with size 10.

\textit{Video}. This model processes raw visual information. For our purposes, we utilise a residual network (ResNet) with 50 layers (He et al. (2016)), which has been widely used in the computer vision community.

\textit{Recurrent Neural Network}. To consider the contextual information in the data, we use RNNs, which can be defined to be either a GRU- or a LSTM-type, and can be used directly to raw information such as physiological data, or to the output of one of the Audio, Video, or fully connected network (FCN) models.

\textit{Fully Connected Network}. A FCN can be defined and used to process either raw input, or the output of one of the Audio, Video, or RNN models.

End2You gives the flexibility to easily combine the aforementioned models. For example, an audio model can be used to process the audio signal and on top a FCN or RNN or both can be stacked. In another example, considering a multimodal case, the video model can be used to process the visual information, and the audio one to process the audio signal. The outputs of these models can be concatenated, and on top can be stacked a RNN, FCN or both. Finally, the user can define her own model and combine it with the predefined models.

\section{System Overview}

\begin{figure*}[h]
\centering
\includegraphics[width=14cm, height=3.5cm]{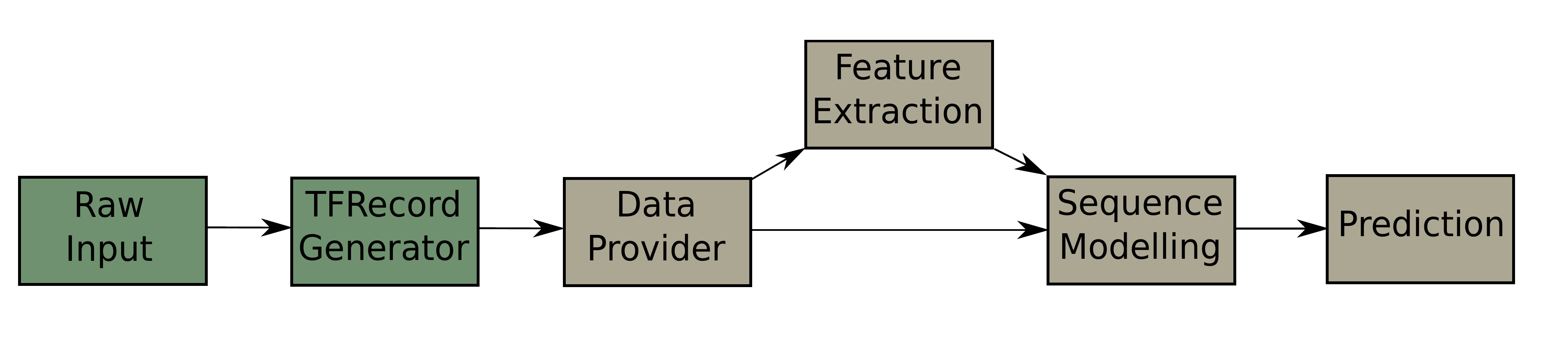}
\caption{The End2You workflow is comprised of the tfrecord generator, the data provider that feeds the either static or dynamic data to the models, and finally, the prediction.}
\label{workflow}
\end{figure*}

End2You is implemented in Python, and can be used either with the Python API or using the command line interface. The code is publicly available on GitHub under a modification of the BSD-3 license\footnote{cf.\ https://github.com/end2you/end2you}. The system workflow, which is shown in Figure~\ref{workflow}, is comprised of two phases. 
First, the raw information needs to be converted to the \textit{.tfrecord} format. Our toolkit provides this capability for both unimodal and multimodal cases. The second phase of our system, which can train or evaluate  a model, is comprised of a data provider that reads the tfrecords and either feeds a feature extraction model (as discussed in the previous section) that extracts features and passes them to sequence models (here RNN), or feeds directly an RNN. The last step is the prediction of the model which is accomplished with the use of FCN. 

\section{Experiments}

To demonstrate the capabilities of End2You, we created both unimodal and multimodal models, and trained them for dimensional affect recognition. To this end, we utilise the REmote COLlaborative and  Affective (RECOLA) database (Ringeval et al. (2013)), used in the Audio/Visual Emotion Challenge and Workshop (AVEC) 2016 (Valstar et al. (2016)). It includes four modalities or sensor-signals, namely, audio,  video,  electro-cardiogram  (ECG),  and  electro-dermal  activity  (EDA). The model needs to predict two time-continuous outputs (arousal and valence).

The unimodal input is comprised of audio, video, or physiological data (ECG), and the multimodal input combines all of them. For the audio and video models, a 2-layer GRU network with 64 units is stacked on top for the prediction. For the physiological modality, the network is a 1-layer GRU with 64 units. The multimodal model is comprised of the video and the audio model, and a FCN of 256 units that combines the features extracted by the unimodal models and also the raw physiological information frames. On top of this model, we stack a 2-layer GRU network. Our loss and metric functions are based on the concordance correlation coefficient (CCC) as was used as competition measure in AVEC.

We compare the results of End2You with the baseline and the winner paper in the AVEC 2016 emotion prediction sub-challenge. Results are depicted in Table~\ref{results}. For all methods four post-processing steps took place on the development set: \textit{(i)} median filtering, \textit{(ii)} centering, \textit{(iii)} scaling, and \textit{(iv)} time-shifting.

Our models perform better than the baseline in both the audio and video modalities, and close for the physiological one. As expected from the literature (Tzirakis et al. (2017)), 
the audio model performs best in the arousal domain and the video in the valence one. Our multimodal model combines the best scores of the unimodal ones. However, this is not enough to surpass the performance of the models from the winner of the challenge and the baseline. We believe that this performance gap is because these models use, among others, physiological hand-crafted features, such as heart rate and heart rate variability (HRHRV), which provide (Valstar et al. (2016))  more information for the emotion recognition task than the raw signals (ECG and EDA).

\begin{table}[ht]
\centering
\begin{tabular}{l|l|r|r} \toprule\midrule
  \textbf{Modality} & \textbf{Method} & \textbf{Arousal} & \textbf{Valence}  \\ \midrule
 \multirow{2}{*}{Audio} & Baseline & .648~(.796) & .375~(.455) \\
 					    & End2You & .669~(.775) & .286~(.382)\\ \midrule
 \multirow{2}{*}{Video} & Baseline & .272~(.379) & .507~(.612) \\
 					    & End2You & .358~(.412) & .561~(.583)\\ \midrule
\multirow{2}{*}{Physio (ECG)} & Baseline & .158~(.271) & .121~(.153) \\
 					          & End2You & .154~(.264) & .052~(.291) \\ \midrule
\multirow{3}{*}{Mutlimodal}  & Baseline & .683~(.821) & .639~(.683) \\ 
 							 & End2You & .672~(.762) & .521~(.493) \\ 
							 & Winner & .770~(.862) & .687~(.750)   \\ 
\end{tabular}
\caption{Results on the test set of the RECOLA database (wrt CCC). In parenthesis are the performances obtained on the development set.} 
\label{results}
\end{table}

\section{Conclusions}

We introduced End2You -- the novel Imperial toolkit for multimodal profiling. It provides capabilities for training and evaluating models with either unimodal or multimodal input. For the multimodal case, the models can be combined in any desirable way. The output of the model can be either for classification or regression tasks, and the number of outputs can be of any value and at any time step, which is defined by the user. Finally, we provide pre-trained unimodal models on the RECOLA database for audio, and visual input.

For future work, we intend to extend our toolkit by introducing models able to handle text information and also additional models for audio and visual analysis will be included. In addition, we will provide an automatic process for face recognition and extraction from videos.

\acks{The authors would like to thank George Trigeorgis for his help in the beginning of this project. The support of the EPSRC Center for Doctoral Training in High Performance Embedded and Distributed Systems (HiPEDS, Grant Reference EP/L016796/1) is gratefully acknowledged.}

\newpage
{\setlength{\parindent}{0cm}
\large{\textbf{References}}
}

\vspace*{5px}
{\setlength{\parindent}{0cm}
J. Chung, C. Gulcehre, K. Cho, and Y. Bengio. Empirical evaluation of gated recurrent
 neural networks on sequence modeling. arXiv preprint arXiv:1412.3555, 2014.
}

\vspace*{5px}
{\setlength{\parindent}{0cm}
N. Dalal and B. Triggs. Histograms of oriented gradients for human detection. In CVPR,
volume 1, pages 886–893, 2005.
}

\vspace*{5px}
{\setlength{\parindent}{0cm}
A. Graves and N. Jaitly. Towards end-to-end speech recognition with recurrent neural
networks. In ICML, pages 1764–1772, 2014.
}

\vspace*{5px}
{\setlength{\parindent}{0cm}
K. He, X. Zhang, S. Ren, and J. Sun. Deep residual learning for image recognition. In
CVPR, pages 770–778, 2016.
}

\vspace*{5px}
{\setlength{\parindent}{0cm}
H. Hermansky. Perceptual linear predictive (plp) analysis of speech. Journal of the Acoustical
Society of America, 87(4):1738–1752, 1990.
}

\vspace*{5px}
{\setlength{\parindent}{0cm}
G. E Hinton, S. Osindero, and Y.-W. Teh. A fast learning algorithm for deep belief nets.
Neural Computation, 18.
S. Hochreiter and J. Schmidhuber. Long short-term memory. Neural Computation, 9(8):
1735–1780, 1997.
}

\vspace*{5px}
{\setlength{\parindent}{0cm}
Y. Kim, H. Lee, and E. M. Provost. Deep learning for robust feature generation in audiovisual
emotion recognition. In ICASSP, pages 3687–3691, 2013.
}

\vspace*{5px}
{\setlength{\parindent}{0cm}
Y. Lecun. Generalization and network design strategies. In Connectionism in perspective.
Elsevier, 1989.
}

\vspace*{5px}
{\setlength{\parindent}{0cm}
B. Logan. Mel frequency cepstral coefficients for music modeling. In ISMIR, volume 270,
pages 1–11, 2000.
}

\vspace*{5px}
{\setlength{\parindent}{0cm}
P. C. Ng and S. Henikoff. Sift: Predicting amino acid changes that affect protein function.
Nucleic acids research, 31(13):3812–3814, 2003.
}

\vspace*{5px}
{\setlength{\parindent}{0cm}
F. Ringeval, A Sonderegger, J. Sauer, and D. Lalanne. Introducing the RECOLA Multimodal
Corpus of Remote Collaborative and Affective Interactions. In FG-Workshop, pages 1–8,
2013.
}

\vspace*{5px}
{\setlength{\parindent}{0cm}
G. Trigeorgis, F. Ringeval, R. Brueckner, E. Marchi, M. A. Nicolaou, B. Schuller, and
S. Zafeiriou. Adieu features? end-to-end speech emotion recognition using a deep
convolutional recurrent network. In ICASSP, pages 5200–5204, 2016.
}

\vspace*{5px}
{\setlength{\parindent}{0cm}
P. Tzirakis, G. Trigeorgis, M. A. Nicolaou, B. Schuller, and S. Zafeiriou. End-to-end
multimodal emotion recognition using deep neural networks. IEEE Journal of Selected
Topics in Signal Processing, 11(8):1301–1309, 2017.
}

\vspace*{5px}
{\setlength{\parindent}{0cm}
M. Valstar, J. Gratch, B. Schuller, F. Ringeval, D. Lalanne, M. Torres Torres, S. Scherer,
G. Stratou, R. Cowie, and M. Pantic. Avec 2016: Depression, mood, and emotion
recognition workshop and challenge. In AVEC, pages 3–10, 2016.
}
\end{document}